\documentclass[smallextended]{svjour3}
\usepackage{amssymb}
\usepackage{amsmath}
\usepackage{graphicx}
\usepackage{adjustbox}
\usepackage[numbers]{natbib}
\usepackage{lineno}
\usepackage{color}
\usepackage{algpseudocode}
\usepackage{algorithm}

\algnewcommand\algorithmicinput{\textbf{Input:}}
\algnewcommand\Input{\item[\algorithmicinput]}
\algnewcommand\algorithmicoutput{\textbf{Output:}}
\algnewcommand\Output{\item[\algorithmicoutput]}

\begin{document}
	
\title{Texture image classification based on a pseudo-parabolic diffusion model \thanks{J. B. Florindo gratefully acknowledges the financial support of S\~ao Paulo Research Foundation (FAPESP) (Grant \#2016/16060-0) and from National Council for Scientific and Technological Development, Brazil (CNPq) (Grants \#301480/2016-8 and \#423292/2018-8). E. Abreu gratefully acknowledges the financial support of São Paulo Research Foundation (FAPESP) (Grant \#2019/20991-8), from National Council for Scientific and Technological Development - Brazil (CNPq) (Grant \#2 306385/2019-8) and PETROBRAS - Brazil (Grant \#2015/00398-0). E. Abreu and J. B. Florindo also gratefully acknowledge the financial support of Red Iberoamericana de Investigadores en Matemáticas Aplicadas a Datos (MathData).}}

\author{Jardel Vieira \and Eduardo Abreu \and Joao B. Florindo}

\institute{
	Jardel Vieira \at 	
	Unidade Acad\^{e}mica Especial de Matem\'{a}tica e Tecnologia - Universidade Federal de Goi\'{a}s\\
	Av. Dr. Lamartine Pinto de Avelar 1120, St. Universit\'{a}rio, 75704-020, Catal\~{a}o, Goi\'{a}s, Brasil.\\
	\email{jardelvieira@ufg.br}	
	\and
	Eduardo Abreu \at 	
	Institute of Mathematics, Statistics and Scientific Computing - University of Campinas\\
	Rua S\'{e}rgio Buarque de Holanda, 651, Cidade Universit\'{a}ria "Zeferino Vaz" - Distr. Bar\~{a}o Geraldo, CEP 13083-859, Campinas, SP, Brasil\\
	\email{eabreu@unicamp.br}		
	\and 
	Joao B. Florindo \at 	
	Institute of Mathematics, Statistics and Scientific Computing - University of Campinas\\
	Rua S\'{e}rgio Buarque de Holanda, 651, Cidade Universit\'{a}ria "Zeferino Vaz" - Distr. Bar\~{a}o Geraldo, CEP 13083-859, Campinas, SP, Brasil\\
	\email{florindo@unicamp.br}		
}

\date{Received: date / Accepted: date}

\maketitle

\begin{abstract}
This work proposes a novel method based on a pseudo-parabolic diffusion process to be employed for texture recognition. The proposed operator is applied over a range of time scales giving rise to a family of images transformed by nonlinear filters. Therefore each of those images are encoded by a local descriptor (we use local binary patterns for that purpose) and they are summarized by a simple histogram, yielding in this way the image feature vector. The proposed approach is tested on the classification of well established benchmark texture databases and on a practical task of plant species recognition. In both cases, it is compared with several state-of-the-art methodologies employed for texture recognition. Our proposal outperforms those methods in terms of classification accuracy, confirming its competitiveness. The good performance can be justified to a large extent by the ability of the pseudo-parabolic operator to smooth possibly noisy details inside homogeneous regions of the image at the same time that it preserves discontinuities that convey critical information for the object description. Such results also confirm that model-based approaches like the proposed one can still be competitive with the omnipresent learning-based approaches, especially when the user does not have access to a powerful computational structure and a large amount of labeled data for training.
\keywords{Pseudo-parabolic equation \and Texture recognition \and Image classification \and Numerical approximation methods for PDEs.}
\end{abstract}

\section{Introduction}

The recognition and classification of texture images, either in isolated conditions or ``in the wild'' is one of the most important tasks in computer vision, with numerous applications in material sciences \cite{NK19}, medicine \cite{JP19}, remote sensing \cite{TKZHLM19}, agriculture \cite{SKSSSRJK19}, etc.

Despite the recent popularization of learning-based approaches like the deep convolutional networks in this area \cite{CMKV16}, there still exists space for the investigation of ``hand-engineered'' image descriptors, especially in situations when there is a small amount of data for training, what is typical, for instance, in medical applications, or when the introduction of specific information from that particular problem (from the physical modeling for example) can contribute with the performance of the computational algorithm.

In this context, and inspired by the recent success of methods like the nonlinear operator derived from a partial differential equation (PDE) in \cite{FB17,F20} , we propose here new texture descriptors based on the application of an operator that corresponds to solutions of a pseudo-parabolic PDE (e.g., \cite{abreu2017computing,AFV20}) when the original image provides the initial condition of that PDE. 
Pseudo-parabolic differential equations are characterized by having mixed time and space
derivatives appearing in the highest-order terms \cite{karch1997asymptotic,ting1969parabolic} and are applied to model many physical phenomena, for instance the fluid dynamic in porous media \cite{hassanizadeh1993thermodynamic}. Here our particular interest on this type of PDE is explained by their ability to regularize a function (image profile in our case) but at the same time that it preserves discontinuities (here corresponding to edges and other relevant local variations important for the image characterization).

To obtain the image representation (descriptors) we follow the PDE application with an encoding process, which here is accomplished by local binary patterns (LBP) \cite{OPM02}. The descriptors are finally submitted to Karhunen-Lo\`{e}ve transform \cite{P1901} and classified by a linear discriminant classifier \cite{F36}. The proposed method is assessed on state-of-the-art benchmark databases (UIUC \cite{LSP05}, UMD \cite{XJF09} and KTHTIPS-2b \cite{HCFE04}) as well as in a practical problem of Brazilian plant species identification \cite{CMB09}. The results confirm that the classification accuracy is competitive both with classical texture descriptors as well as with modern approaches like convolutional neural networks. 

This text is divided into 7 sections, including this introduction. In the next one, we present a brief review of other PDE-based methods for image analysis and processing in the literature. In Section 3, the PDE model is presented together with the numerical scheme used here to solve it. Section 4 shows the methodology, including a short introduction to LBP theory and the proposed method. Section 5 describes the setup for the texture classification experiments. Section 6 presents the results of these experiments in comparison with the state-of-the-art on this topic and corresponding discussion. Finally, Section 7 concludes the work, summarizing the main points and raising some general discussion on the impact of the proposed methodology.

\section{Related works}

Most methods for the classification of texture images can be divided into two main categories: traditional approaches and deep learning methods \cite{LCFZCP19}. In the first group, we have the classical second order statistical methods like Haralick descriptors \cite{H73} and Local Binary Patterns (LBP) \cite{OPM02}, more recently we have seen the Scale-Invariant Feature Transform (SIFT) \cite{L04}, Filter Banks \cite{VZ09}, Fisher Vector \cite{PSM10}, Fractal Descriptors \cite{FB16}, PCANet \cite{CJGLZM15}, Locally Encoded Transform Feature Histogram \cite{SLMWC18}, and many others. On the second category, deep learning methods, we have those based on Convolutional Neural Networks (CNN), such as Deep Convolutional Activation Feature (DeCAF) \cite{DJVHZTD14}, Deep Filter Banks \cite{CMKV16}, Deep Texture Encoding Network (DeepTEN) \cite{ZXD17}, Locally Transfered Fisher Vectors (LFV) \cite{SZLHOC17}, Deep Texture Manifold (DEP) \cite{XZD18}, First and Second Order Network (FASON) \cite{DND17}, Multiple-Attribute-Perceived (MAP-Net) \cite{ZCZZ19}, and others.

The disseminated use of PDEs to provide different viewpoints of a digital image can be traced back to the space-scale theory of Witkin \cite{W83}, formulated in terms of the canonical model of diffusion equation, i.e., the heat equation. The well-known maximum principle of the elliptic equations was employed in that occasion to ensure that spurious details not present in finer scales would also not appear in coarser scales. This is a fundamental property that must be satisfied by any robust multiscale theory \cite{K84}.

However the smoothing effect intrinsic to diffusion equations can also represent a problem in image analysis as at the same time that the control over the smoothing parameter allows for fine tuning of the desired scale of observation, it also smooths out image edges. Edges are known to be powerful attributes in the discrimination of objects present in the image. To address this issue, Perona and Malik proposed the anisotropic diffusion equation in \cite{PM90}. This ensures the smoothing multiscale effect while preserving edges by means of a less aggressive action of the operator in regions with high-valued gradients.

Evolutions over the anisotropic model can be found in \cite{CLMC92}, where the image is regularized previous to the application of the PDE operator. \cite{CG93} and \cite{W96} employ diffusion tensors to address the problem of ``ghost'' stair-case edges arising in some situations after the application of the operator. A review of this family of approaches can be found in \cite{W97}. Another strategy, focused especially in speckle reduction and avoiding the direct use of gradients, is presented in \cite{YA02}. \cite{GKL13} and \cite{G09} also propose an interesting alternative named Backward-Forward Regularized Diffusion, again attenuating the ``stair-case'' effect and using two regularizing parameters. More recently, a promising method combining the model in \cite{GKL13} with texture descriptors is also proposed in \cite{NGB18}.

With regards to pseudo-parabolic equations, the early existence, uniqueness, and regularity theory is well developed, for example, in \cite{showalter1969partial,showalter1975nonlinear,showalter1970pseudoparabolic,ting1969parabolic}.
Such theory predicts that the additional pseudo-parabolic term decreases the smoothing property characteristic to parabolic problems.
This character has consequences on the behavior of the solution and it was also observed 
in subsequent works.
For instance, if the initial data has jump discontinuity at some point, then so does the solution for every time \cite{cuesta2003model,cuesta2009numerical}. Here this is a particularly important feature as it allows a controlled smoothing effect preserving relevant edges to some extent.
It is also worth to mention that, in general, there is no maximum principle for pseudo-parabolic equations 
such as one expected of solutions to parabolic equations \cite{stecher1977maximum}, although the specific simplified model adopted here still preserves that principle.

\section{Proposed methodology}

\subsection{Partial differential equation and numerical modeling}\label{sec:pde}

The model of partial differential equation adopted here is that of a pseudo-parabolic type. In its more general form, those equations can be represented by
\begin{equation}
\frac{\partial}{\partial t}\mathcal{A}(u) + \mathcal{B}(u) = 0,
\end{equation}
where $\mathcal{A}$ and $\mathcal{B}$ are nonlinear elliptic operators. More specifically, we build our model on the pseudo-parabolic equation:
\begin{equation}
\frac{\partial u}{\partial t} 
= 
\nabla \cdot  
\left[ g(x , y , t) \, \nabla \left(u
+ \tau \,  \frac{\partial u}{\partial t}  \right) \right].
\label{eq:pptransp2}
\end{equation}
We also identify the pseudo-parabolic diffusive flux $\mathbf{w}$ as given by
\begin{equation}
\mathbf{w} =   
g(x,y,t) \, \nabla \left(u 
+ \tau \,  \frac{\partial u}{\partial t} \right).
\end{equation}

Let $\Omega \subset \mathbb{R}^{2}$ denote a rectangular domain and ${\displaystyle u( \,\cdot\,,\, \cdot \, , \,t):\Omega \rightarrow \mathbb {R} }$  be a family of gray scale images that satisfies the pseudo-parabolic equation \eqref{eq:pptransp2}. 
The original image  corresponds to $t = 0$, i.e., an initial value for the differential problem.
To completely define the differential problem, we impose zero flux condition across the domain boundary $\partial \Omega$,
\begin{equation}
\mathbf{w} \cdot \mathbf{n} = 0, \qquad (x \,,\, y) \in \partial \Omega.
\end{equation}

In general, the coefficient $g(x \,,\, y \,,\, t)$ is a positive definite tensor and,
in the context of nonlinear diffusion models for image processing, it may depend on $\nabla u$ \cite{PM90,CLMC92,W97}. 
In this work, we describe the numerical approach considering $g$ as a scalar. 
However, the method may be directly extended for the others cases.

For the numerical modeling we employ a uniform rectangular mesh and cell-centered finite
differences.
To construct this scheme, we use ideas learned from the mixed
finite element method in \cite{abreu2017computing,AFV20}.
The mixed finite element framework has two advantages \cite{chavent1991unified}:
it yields a very natural and physical discretization
of the boundary conditions; 
and	it gives a consistent way of defining a gradient flux, as in
the mixed formulation.

Now, we will discuss a finite difference approach for the pseudo-parabolic
equation \eqref{eq:pptransp2}.
Consider a uniform partition of $\Omega$ into rectangular subdomains
$\Omega_{i,j}$, for $i = 1 \,,\, \dots \,,\, m$ and $j = 1 \,,\, \dots \,,\, l$, with dimensions 
$\Delta x \times \Delta y$. 
The center of each subdomain $\Omega_{i,j}$ is denoted by $(x_i \,,\, y_j)$.
Given a final time of simulation $T$, consider a uniform partition of
the interval $[0 \,,\, T]$ into $N$ subintervals.
The time step $\Delta t = T/N$  is usually defined by a stability
condition.
We denote the time instants as $t_n = n \Delta t$, 
for $n = 0 \,,\, \dots \,,\, N$.

Let $U_{i,j}^n$ be a finite difference approximation for 
$u(x_i \,,\, y_j \,,\, t_n)$. 
A discretization of \eqref{eq:pptransp2} by the finite difference
method is given by
\begin{equation}
\frac{U_{i,j}^{n+1} - U_{i,j}^n}{\Delta t}
=
\frac{W_{i+\frac{1}{2},j}^{n+1} - W_{i-\frac{1}{2},j}^{n+1}}{\Delta x}
+ \frac{W_{i,j+\frac{1}{2}}^{n+1} - W_{i,j-\frac{1}{2}}^{n+1}}{\Delta y},
\end{equation}
where the approximation of the diffusive flux is given by a 
centered difference formula:
\begin{subequations}
	\begin{equation}
	W_{i+\frac{1}{2},j}^{n+1} =
	G_{i+\frac{1}{2},j}^n \left( \frac{U_{i+1,j}^{n+1} - U_{i,j}^{n+1}}{\Delta x} \right) 
	+ \frac{\tau \, G_{i+\frac{1}{2},j}^n}{\Delta t} \left( 
	\frac{U_{i+1,j}^{n+1} - U_{i,j}^{n+1}}{\Delta x}
	- \frac{U_{i+1,j}^{n} - U_{i,j}^{n}}{\Delta x}
	\right),
	\end{equation}
	\begin{equation}
	W_{i,j+\frac{1}{2}}^{n+1} =
	G_{i,j+\frac{1}{2}}^n \left( 
	\frac{U_{i,j+1}^{n+1} - U_{i,j}^{n+1}}{\Delta y}
	\right) 
	+ \frac{\tau \, G_{i,j+\frac{1}{2}}^n}{\Delta t} \left( 
	\frac{U_{i,j+1}^{n+1} - U_{i,j}^{n+1}}{\Delta y}
	- \frac{U_{i,j+1}^{n} - U_{i,j}^{n}}{\Delta y}
	\right).
	\end{equation}
\end{subequations}

The coefficients  are chosen as the arithmetic mean at interfaces:
\begin{equation}
G_{i+\frac{1}{2},j}^n = 
\frac{G_{i,j}^n + G_{i+1,j}^n}{2},
\qquad
G_{i,j+\frac{1}{2}} = 
\frac{G_{i,j}^n + G_{i,j+1}^n}{2},
\end{equation}
where $G_{i,j}^n$ is an approximation for $g(x_i \,,\, y_j \,,\, t_n)$.

We chose to use the coefficients at time $t_n$.
This choice leads to an algebraic linear system over
the unknowns $U_{i,j}^{n+1}$.
The algebraic equations may be written as
\begin{equation}
a_{i,j-\frac{1}{2}} \, U_{i,j-1}^{n+1}  
+ a_{i-\frac{1}{2},j} \, U_{i-1,j}^{n+1} 
+ a_{i,j} \, U_{i}^{n+1} 
+ a_{i+\frac{1}{2},j} \, U_{i+1,j}^{n+1}
+ a_{i,j+\frac{1}{2}} \, U_{i,j+1}^{n+1} =
b_{i,j},
\end{equation}
where,
\begin{equation}
\begin{array}{ll}
b_{i,j} = 
U_{i,j}^{n}
-\tau  &\left[\, \dfrac{
	G_{i-\frac{1}{2},j}^n \,  U_{i-1,j}^n - ( G_{i-\frac{1}{2},j}^n + G_{i+\frac{1}{2},j}^n) U_{i,j}^n + G_{i+\frac{1}{2},j}^n \, U_{i+1,j}^n
}{\Delta x^2} \right.
\\
&\left.
+  \,\dfrac{
	G_{i,j-\frac{1}{2}}^n \, U_{i,j-1}^n  
	-( G_{i,j-\frac{1}{2}}^n +  G_{i,j+\frac{1}{2}}^n) 
	U_{i,j}^n 
	+ G_{i,j+\frac{1}{2}}^n \, U_{i,j+1}^n
}{\Delta y^2} \right],
\end{array}
\label{eq:rhs}
\end{equation}
and the coefficients are given by
\begin{subequations}
	\begin{equation}
	a_{i,j-\frac{1}{2}} = 
	- \frac{(\Delta t  + \tau)}{\Delta y^2} \, G_{i,j-\frac{1}{2}}^n, 
	\qquad
	a_{i,j+\frac{1}{2}} = 
	- \frac{(\Delta t + \tau)}{\Delta y^2} \, G_{i,j+\frac{1}{2}}^n, 
	\end{equation}
	\vspace{-0.3cm}
	\begin{equation}
	a_{i-\frac{1}{2},j} = 
	- \frac{ (\Delta t  + \tau)}{\Delta x^2} \, G_{i-\frac{1}{2},j}^n, 
	\qquad
	a_{i+\frac{1}{2},j} = 
	- \frac{(\Delta t + \tau)}{\Delta x^2} \, G_{i+\frac{1}{2},j}^n,
	\end{equation}
	\vspace{-0.3cm}
	\begin{equation}
	a_{i,j} = 
	-(a_{i,j-\frac{1}{2}} + a_{i,j+\frac{1}{2}} + a_{i-\frac{1}{2},j} + a_{i+\frac{1}{2}},j) + 1
	\end{equation}
	\label{eq:matcoef}
\end{subequations}

Finally, to impose the boundary conditions, we have written the algebraic equations for the subdomains that intersect the boundary.

For each time step, we have to solve a linear system 
$\mathbf{A}^n \, \mathbf{U}^{n+1} = \mathbf{b}^n$, where $\mathbf{U}^{n+1}$ is the vector of unknowns (numerical solution approximation at $t_{n+1}$) and $\mathbf{b}^n$ is the vector of right-hand side terms \eqref{eq:rhs}. 
The  matrix $\mathbf{A}^n$ of linear system  is symmetric positive definite and sparse, thus efficient methods may be applied to solve the algebraic system.
In this work, we use the Preconditioned Conjugate Gradient Method (PCG).

In this work, we consider the $g(x \,,\ y \,,\, t) \equiv 1$, thus Equation \eqref{eq:pptransp2} becomes a linear pseudo-parabolic equation.
For an image to be processed, each subdomain $\Omega_{i,j}$ corresponds to a pixel, i.e., the mesh parameters are $\Delta x = \Delta y = 1$.
We choose the time step $\Delta t = \Delta x$ and the damping coefficient $\tau = 5$.

Algorithms \ref{alg:pp} and \ref{alg:ls} shows the computational routines involved in the described numerical scheme in a pseudo-code language. Algorithm \ref{alg:ls} refers to the inner subdomains. Those sudomains intercepting the boundary can be implemented similarly with only small adaptations. More details can be found in \cite{VA18}.

\begin{algorithm}[H]
	\caption{Algorithm.}
	\label{alg:pp}
	\begin{algorithmic}[1]
		\Input $U^0$
		\Output $U^n$
		\For{$n = 1 \textrm{ to } N$}
		\State $[A , b] \gets \mathrm{LINSYST}(U^{n-1})$
		\State $U^{n} \gets \mathrm{PCG}(A, b, U^{n-1})$ 
		\EndFor
	\end{algorithmic} 
\end{algorithm}

\begin{algorithm}[H]
	\caption{Linear system.}
	\label{alg:ls}
	\begin{algorithmic}[1]
		\Function{LINSYST}{$G$,$U$}
		
		\State $[m,l] \gets \mathrm{size}(U)$
		
		\State $A \gets \mathrm{zeros}(m \cdot l \,,\, m \cdot l)$
		\State $b \gets \mathrm{zeros}(m \cdot l \,,\, 1)$
		
		\For{$j = 2 \textrm{ to } l-1$}
		\For{$i = 2 \textrm{ to } m-1$}
		\State $k = (j-1) \cdot m + i$
		\State $A_{k,k} \gets 4 \, (1 + \tau) + 1$
		\State $A_{k,k-m} \gets - (1  + \tau)$
		\State $A_{k,k-1} \gets - (1  + \tau)$
		\State $A_{k,k+1} \gets - (1  + \tau)$
		\State $A_{k,k+m} \gets - (1  + \tau)$
		\State $ b_k \gets U_{i,j} - \tau \,
		[U_{i,j-1} + U_{i-1,j} - 4 \, U_{i,j} + U_{i+1,j} + U_{i,j+1}]
		$
		\EndFor 
		\EndFor
		
		\State \Return $[A,b]$
		\EndFunction
	\end{algorithmic} 
\end{algorithm}

\begin{table}[h]
	\centering
	\caption{Variables used in Algorithm \ref{alg:pp} and \ref{alg:ls}.}	
	\begin{tabular}{ll}
		\hline
		$U^0$ & Matrix $ m \times l $ containing the texture image in gray scales\\
		$U^n$ & \\
		$A$ & Matrix $ m \cdot l \times m \cdot l $ for algebraic linear system\\
		$b$ & Array $ m \cdot l \times 1 $ containing independent terms of linear system\\		
		\hline
	\end{tabular}
	\label{tab:variables}		
\end{table}

\begin{table}[h]	
	\centering
	\caption{Pre-programmed routines supposed to be available by Algorithm \ref{alg:pp} and \ref{alg:ls}.}	
	\begin{tabular}{ll}
		\hline
		$\mathrm{size}(U)$ & Return the size (number of rows and columns) of the matrix $A$ \\
		$\mathrm{PCG}(A, b, U)$ & Return the solution of the linear system $Ax=b$ by the PCG method with \\ &initial approximation $U$\\	
		\hline
	\end{tabular}
	\label{tab:routines}	
\end{table}

\subsection{Local binary patterns}

Local binary patterns (LBP) \cite{OPM02} are gray-level texture descriptors, developed mainly for texture classification purposes. In its original and most simple version, LBP assigns a code to each pixel (central pixel) taking into consi\-deration its gray value $g_c$ and the gray values $g_p$ of $P$ neighbor pixels equally spaced on a circle centered at the central pixel and with radius $R$. Such code is provided by
\begin{equation}
	\label{eq:lbp}
	LBP_{P,R} = \sum_{p=0}^{P-1}H(g_p-g_c)2^p,
\end{equation}
where $H$ is the step (Heaviside) function: $H(x) = 1$ if $x \geq 0$ and $H(x) = 0$, otherwise, and the coordinates of $g_p$ are given by
\begin{equation}
	(x_p,y_p) = (x_c - R\sin(2\pi p/P),y_c + R\cos(2\pi p/P)),
\end{equation}
being $(x_c,y_c)$ the coordinates of the central pixel. If $(x_p,y_p)$ are not integer values, $g_p$ value is obtained by linear interpolation.

Here we use an improved version of (\ref{eq:lbp}), which ensures rotation invariance and is more discriminative. It discards patterns that are very similar, the so-called uniform patterns. The new descriptor is given by
\begin{equation}
	LBP_{P,R}^{riu2} = \left\{
		\begin{array}{ll}
			\sum_{p=0}^{P-1}H(g_p-g_c)2^p & \mbox{if } \mathcal{U}(LBP_{P,R})\geq 2\\
			P+1 & \mbox{otherwise}, 
		\end{array}
	\right.
\end{equation}
provided that
\begin{equation}
	\mathcal{U}(LBP_{P,R}) = |H(g_{P-1}-g_c)-H(g_0-g_c)| + \sum_{p=1}^{P-1}|H(g_{p}-g_c)-H(g_{p-1}-g_c)|.
\end{equation}

\subsection{Proposed descriptors}

Recalling the notation in Section \ref{sec:pde}, the texture descriptors developed here are obtained by evolving the pseudo-parabolic operator on the image $u(\cdot,\cdot,t_0)$ over a range of time $t_1,t_2,\cdots,t_N$. In this way we have a family of transformed images $\{u(\cdot,\cdot,t_n)\}$ for $n=1,\cdots,N$. Empirically we found out that $N=50$ provide a reasonable compromise between computational cost and recognition accuracy.

In the next step, we apply the $LBP_{P,R}^{riu2}$ operator over each evolved image. We adopted the combinations of $P,R$ values listed in Table \ref{tab:PR}.
\begin{table}[!htpb]
	\centering
	\caption{Parameters of $LBP_{P,R}^{riu2}$ codes used in this work.}\label{tab:PR}	
	\begin{tabular}{c|cccc}
		P & 1 & 2 & 3 & 4\\
		\hline
		R & 8 & 16 & 24 & 24\\
	\end{tabular}
\end{table}

These combinations were chosen both based on recent publications on LBP, e.g. \cite{LFGWP17}, which demonstrated their suitability, and on empirical tests specific for the problem that we are addressing here.

Finally, as usual in the LBP pipeline, we compute the histogram $\mathrm{h}$ for all possible values of $LBP_{P,R}^{riu2}$ codes. We can formally summarize the descriptors by
\begin{equation}
	\mathfrak{D}(u) = \bigcup_{\substack{(P,R) = \{(8,1),\\(16,2),(24,3),\\(24,4)\}}}\bigcup_{n=0}^N \mathrm{h}(LBP_{P,R}^{riu2}(u(\cdot,\cdot,t_n))).
\end{equation}

\begin{figure}[!htpb]
	\centering
	\includegraphics[width=\textwidth]{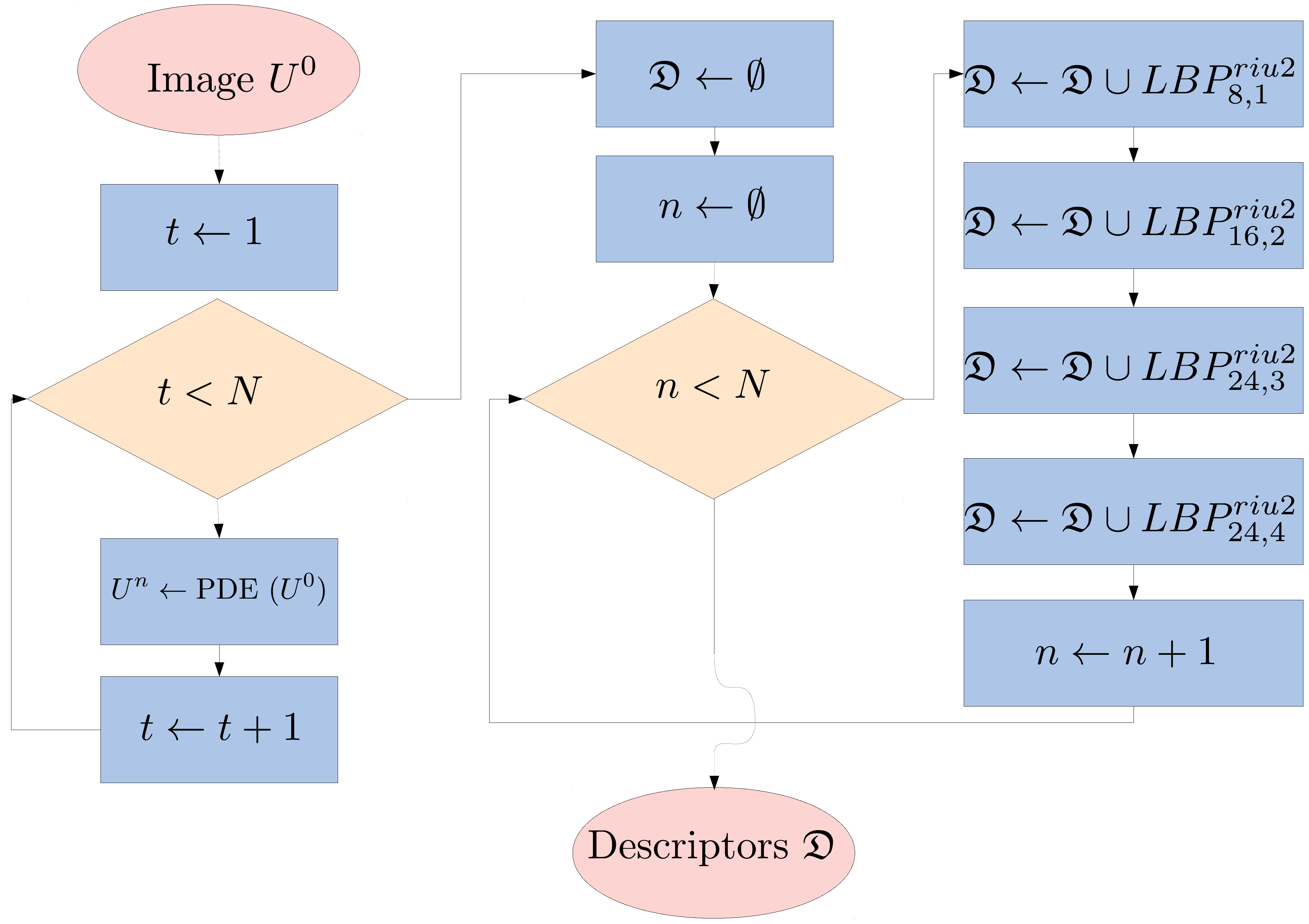}
	\caption{Flow chart of the proposed method.}
	\label{fig:method1}
\end{figure}

\begin{figure}[!htpb]
	\centering
	\includegraphics[width=\textwidth]{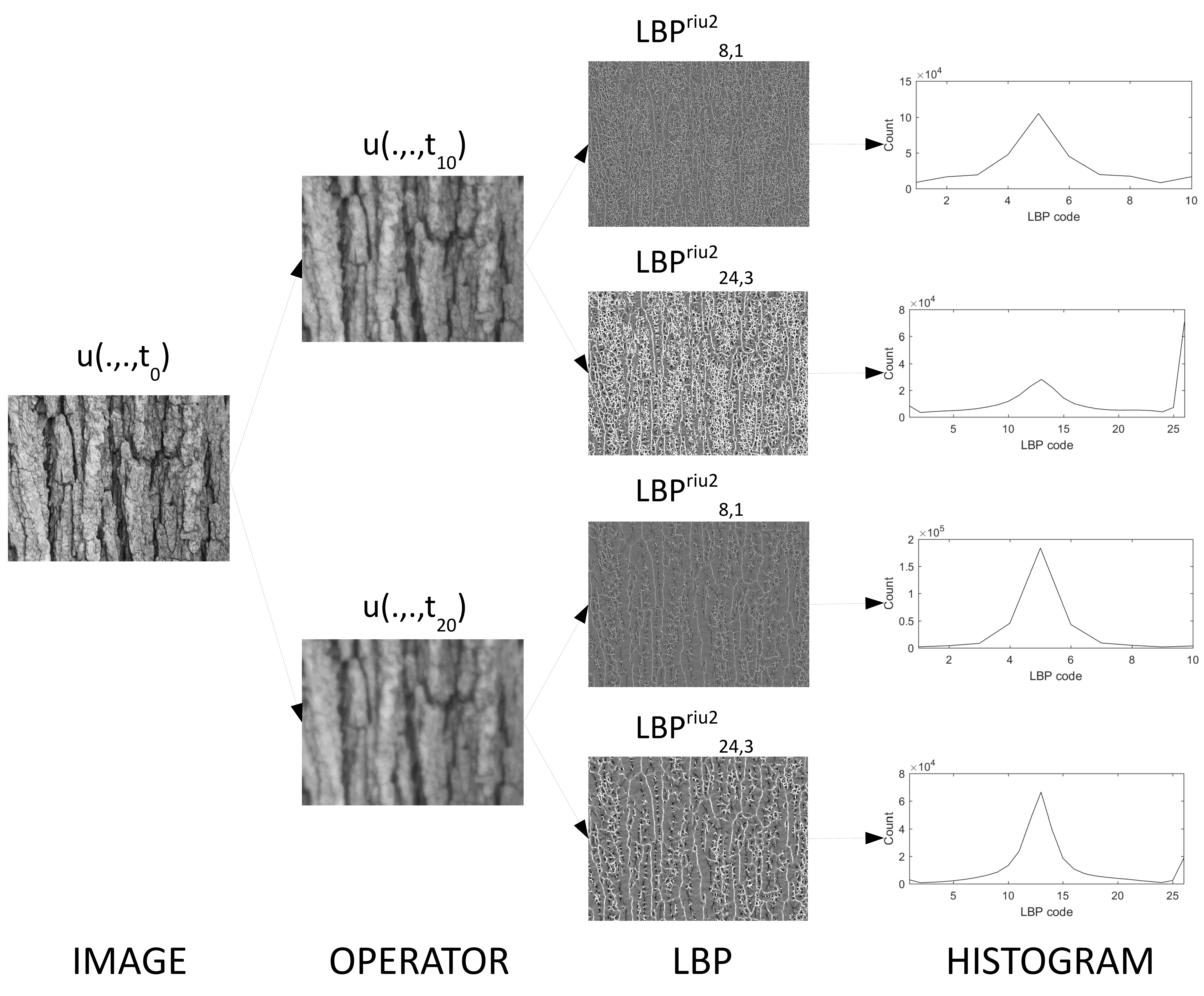}
	\caption{Outcomes of the most important steps of the proposed descriptors: original texture image, application of the pseudo-parabolic operator, LBP encoding and histogram.}
	\label{fig:method}
\end{figure}

We end up with a large feature vector for each image, with 4000 components in each one. This requires some strategy to avoid issues like the dimensionality curse. To cope with that we apply Karhunen-Lo\`{e}ve transform \cite{P1901} and select the number of uncorrelated features that provides sufficiently discriminative capacity. Figure \ref{fig:method1} shows a flow chart of the main steps involved in the proposed method, whereas Figure \ref{fig:method} visually illustrates the outcomes of those correspon\-ding steps for an exemplar texture image. To facilitate the visualization, we show only some of the parameter values ($t_n$, $P$ and $R$) used to provide the descriptors.

\subsection{Motivation}

The numerical model employed here is inspired by works like \cite{van2007new}, where the the Buckley-Leverett equation is extended to a two-phase flow in porous media with dynamic capillary pressure, yielding the following general equation:
\begin{equation}
\frac{\partial \,}{\partial t}(\phi u) +
\frac{\partial F(u)}{\partial u} = 
\frac{\partial }{\partial x} \left( G(u) 
\frac{\partial}{\partial x} \left( J(u) 
+ \tau \frac{\partial \,}{\partial t}(\phi u) \right) \right).
\label{PPE}
\end{equation}
This model presents some important characteristics that are helpful for an application like the one described here. First, it is sufficiently flexible to represent a diversity of physical phenomena, partially due to the existence of a third order mixed derivative term. It can be demonstrated, for example, that the coefficient $\tau$ is critical for the type of solution profile with respect to the regularized Buckley-Leverett problem. Such parameter acts as a bifurcation threshold: if $\tau$ is within a certain range, the solution profile is similar to the classical shock waves; on the other hand, for higher values of $\tau$, the profile changes abruptly and new types of shock waves take place in the solution. In some extreme cases, the solution may contain a non-monotone profile or even exhibit damped oscillation.

Another interesting characteristic is that pseudo-parabolic equations like that in \cite{van2007new} can also be interpreted as a regularization of the hyperbolic Buckley-Leverett equation \cite{abreu2017computing,AFV20}. Nevertheless, unlike classical regularization schemes based, for example, on standard diffusion processes, the pseudo-parabolic term attenuates the smoothing effect typically associated to parabolic solutions. It is known, for instance, that jump discontinuities present at some time persist in the solution for the remaining of the time evolution. In practice, this implies that the model employed here is strongly characterized by regularity preservation. Whereas parabolic differential equations are known by their effect of increasing regularity regardless the information conveyed by the image, here the edges and other important elements of the image expressed by discontinuities are not vani\-shed by the operator with the same aggressiveness. Preserving discontinuities is an important feature in space-scale approaches to image analysis and well explored, for example, in the classical work of Perona and Malik \cite{PM90}.

Generally speaking, the numerical modeling employed here consists of a fully coupled space-time approach, capable of appropriately accounting for the diffusive flux naturally appearing in the pseudo-parabolic equation. In terms of its use as an image analysis tool, it implies in controlled conditions for the regularization process. Being followed by the local encoding of binary patterns makes the proposed descriptors robust in many aspects both to geometrical variations addressed by LBP as well as to edge localization preserved by the nonlinear differential operator.

\section{Experiments}

The accuracy of the proposed method in texture classification is assessed over three benchmark data sets and compared with other approaches, some of them classically used for this purpose, other ones from the state-of-the-art in the literature.

UIUC \cite{LSP05} is a data set of gray-level texture images. Each image has dimensions 256$\times$256 and we have 40 images per class and 25 classes, totalizing 1000 images. Those images are collected under uncontrolled conditions with variations in scale, illumination, perspective and albedo. The training/testing split follows the protocol classically used with half of the images from each class randomly selected for training and the remaining ones for testing. Such random division is repeated 10 times to compute the average accuracy of the classification process.

UMD \cite{XJF09} is another gray-scale texture database that shares some similarities with UIUC, like the number of classes and images per class. The images are also acquired under uncontrolled conditions. Nevertheless, UMD images have high resolution, each sample has dimensions 1280$\times$960. The intra-class variation in scale and viewpoint is also more intense, which adds difficulties to the class-wise discrimination. The training/testing split follows the same protocol of UIUC.  

Finally, KTHTIPS-2b \cite{HCFE04} is a collection if color texture images (here we convert the samples to gray-scales), comprising a total of 4752 images divided into 11 classes, each one corresponding to a particular material, like wood, plastic, bread, etc. The most important motivation behind the development of this database is the focus on the real material regardless the conditions under which the object was photographed. Each class is divided into 4 samples, and each sample represents a particular setting of illumination, scale and pose. The training/testing split is the classically used in the literature, where one sample is used for training and the remaining ones are employed for testing.

The feature vectors resulting from the concatenation of different transform parameters and LBP settings are large and a Karhunen-Lo\'{e}ve is used to reduce dimensionality. Finally, we tested three classifiers to process the descriptors, i.e., linear discriminant analysis (LDA) \cite{F36}, support vector machines (SVM) \cite{CV95}, and random forests (RF) \cite{H95}.

\section{Results and Discussion}

\begin{figure}[!htpb]
	\centering
	\includegraphics[width=.8\textwidth]{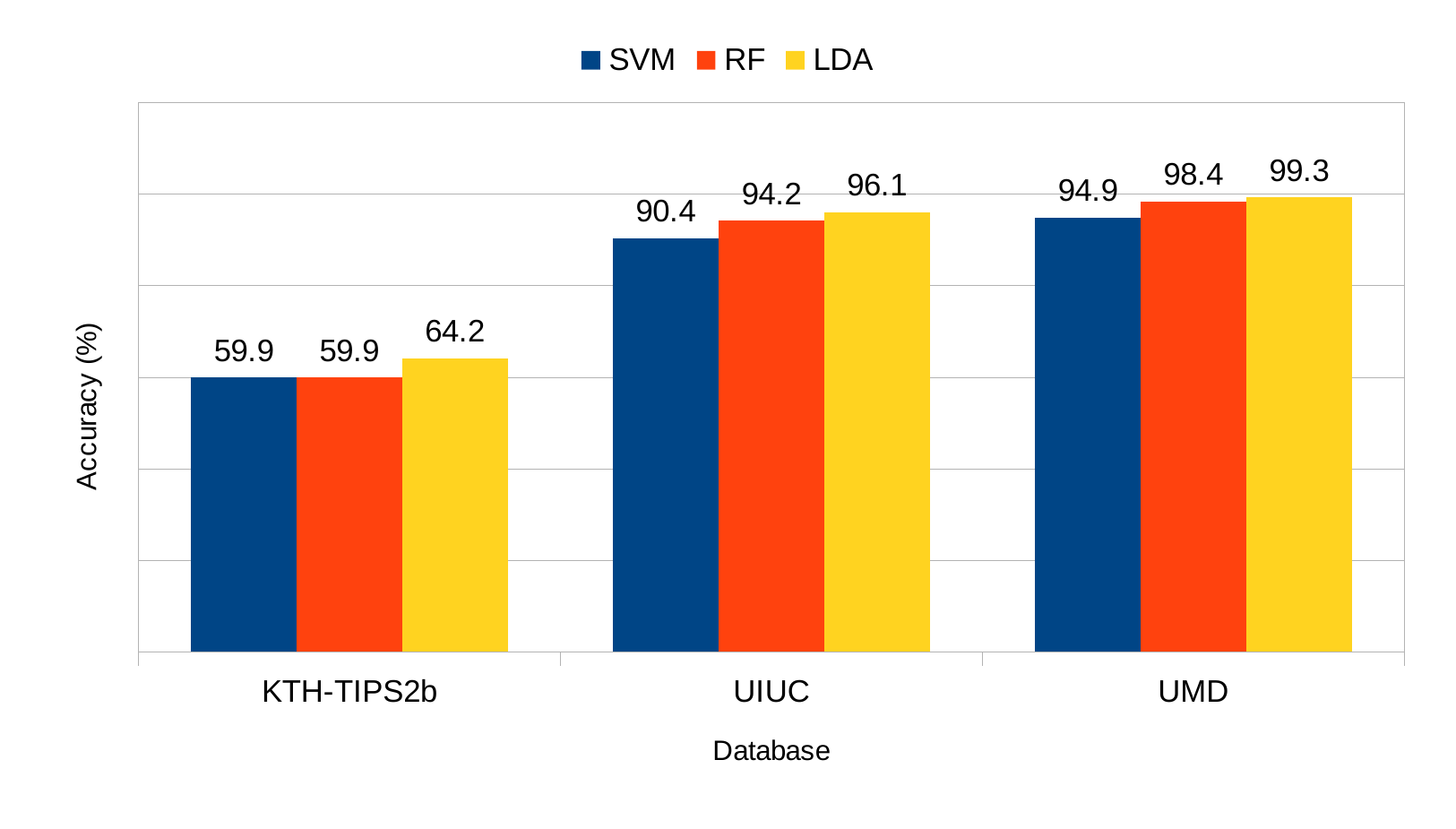}
	\caption{Classification accuracy using different classifiers.}
	\label{fig:classif}
\end{figure}

Figure \ref{fig:classif} shows the classification accuracy when using LDA, SVM and RF. All these classifiers use the data in its original form, without any transformation. In this way they are expected to highlight the discriminative power of the descriptors. LDA outperforms SVM and RF in this task with relevant difference. The nature of each classifier can explain such difference. LDA was originally developed to handle multi-class problems, avoiding strategies like ``one-against all'' that need to be employed by SVM. Especially in ideal situations when the classes are balanced, methods dealing with all classes on an equal footing tend to work more appropriately.

\begin{figure}[!htpb]
	\centering
	\begin{tabular}{cc}
		\includegraphics[width=.45\textwidth]{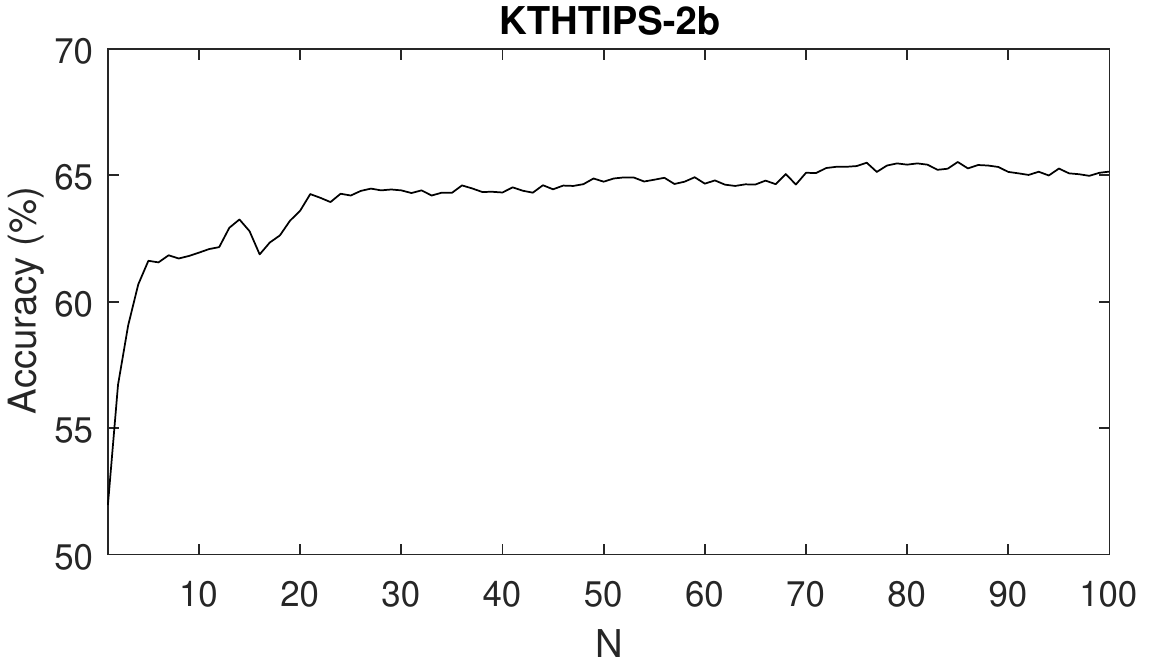} &
		\includegraphics[width=.45\textwidth]{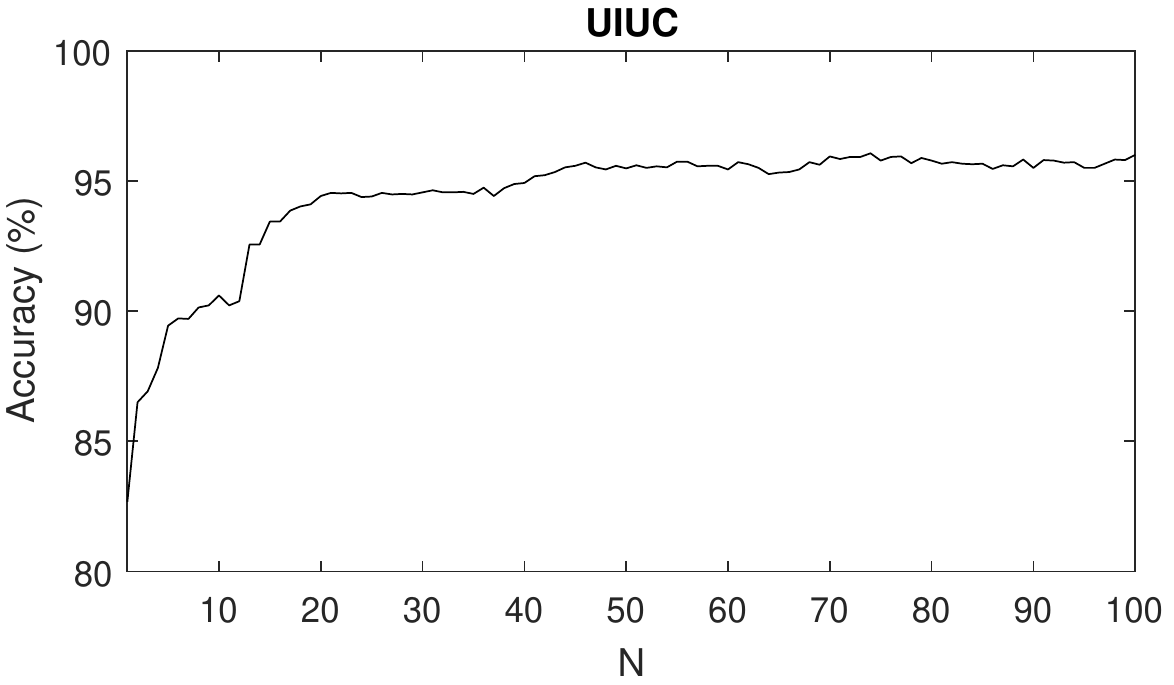}\\
	\end{tabular}
	\begin{tabular}{c}
		\includegraphics[width=.45\textwidth]{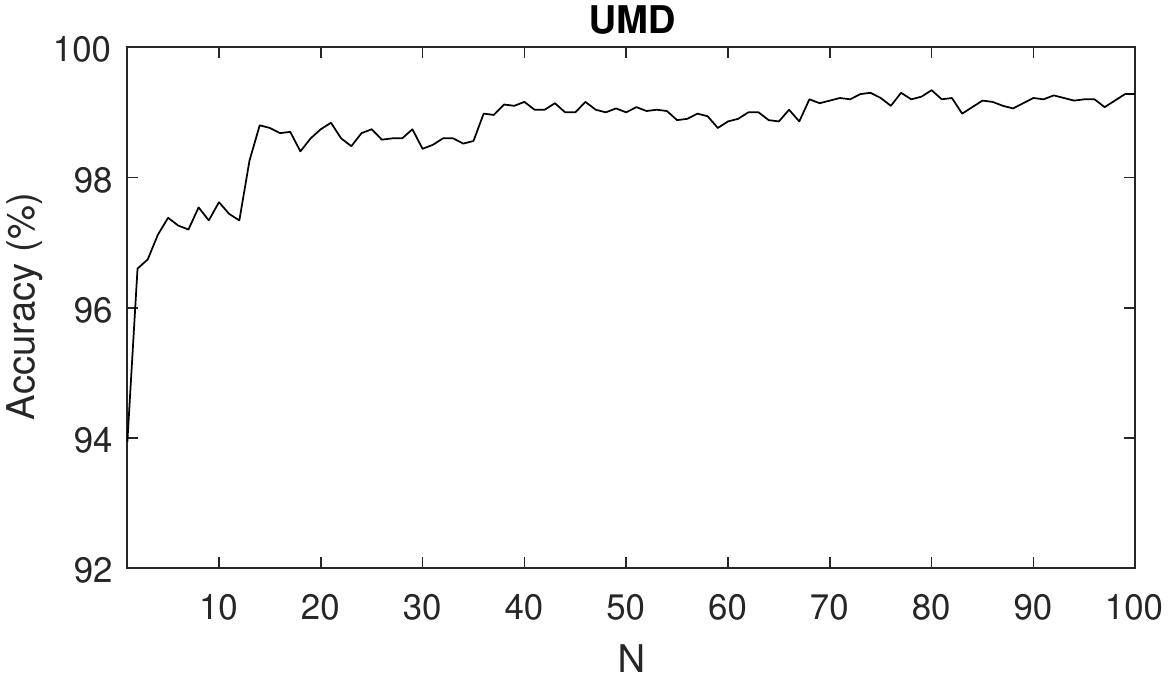}\\
	\end{tabular}
	\caption{Accuracy for different values of $N$.}
	\label{fig:N}
\end{figure}

Figure \ref{fig:N} depicts the behavior of the classification accuracy when the parameter $N$ (maximum iteration time of the PDE operator) is varied. This is actually the most relevant free parameter that we have to tune in this method. According to the plots, the results achieved by $N\geq 50$ are quite similar. This is the critical point where the image is so smoothed  that no more relevant information can be captured by the image descriptors. Based on this outcome, we employ $N=50$ for the remaining experiments.

Table \ref{tab:SRdatabase} lists the accuracy (percentage of images assigned by the classifier to the correct classes) for each database and compared with results recently published in the literature on the same databases. Our approach outperforms several ``handcrafted'' descriptors and even some learning-based methods like those described in \cite{CMKV16}. Whereas in UIUC and UMD several solutions have already been proposed that achieve accuracy close to 100\%, KTHTIPS-2b is much more challenging, especially due to its training protocol, in which the algorithm should be able to recognize a material at different viewpoints, illuminations, etc. based on features representing only the material at an original condition. Even here, our proposal outperforms several ``hand-engineered'' descriptors like SIFT and LBP and was also competitive with the modern CNN-based approaches.
\begin{table}[!htpb]
	\centering
	\caption{Accuracy of the proposed descriptors compared with other texture descriptors in the literature. A superscript $^1$ in KTHTIPS-2b means training on three samples  and testing on the remainder.}
	\label{tab:SRdatabase}
	\resizebox{\textwidth}{!}{
	\begin{tabular}{cc}
		KTHTIPS-2b & UIUC\\
		\begin{tabular}{cc}
			\hline
			Method & Acc. (\%)\\
			\hline
			VZ-MR8 \cite{VZ05} & 46.3\\
			LBP \cite{OPM02} & 50.5\\
			VZ-Joint \cite{VZ09} & 53.3\\
           	BSIF \cite{KR12} & 54.3\\			
			LBP-FH \cite{AMHP09} & 54.6\\
			CLBP \cite{GZZ10} & 57.3\\
           	SIFT+LLC \cite{CMKV16} & 57.6\\         			
			ELBP \cite{LZLKF12} & 58.1\\
			SIFT + KCB \cite{CMKMV14} & 58.3\\
			SIFT + BoVW \cite{CMKMV14} & 58.4\\
           	LBP$_{riu2}$/VAR \cite{OPM02} & 58.5\footnotemark[1]\\			
           	PCANet (NNC) \cite{CJGLZM15} & 59.4\footnotemark[1]\\			
           	RandNet (NNC) \cite{CJGLZM15} & 60.7\footnotemark[1]\\			
           	ScatNet (NNC) \cite{BM13} & 63.7\footnotemark[1]\\
			\hline
			Proposed & 65.5\\
			\hline
		\end{tabular}
		&
		\begin{tabular}{cc}
			\hline
			Method & Acc. (\%)\\
			\hline
			RandNet (NNC) \cite{CJGLZM15} & 56.6\\
			PCANet (NNC) \cite{CJGLZM15} & 57.7\\
			BSIF \cite{KR12} & 73.4\\
			VZ-Joint \cite{VZ09} & 78.4\\
			LBP$_{riu2}$/VAR \cite{OPM02} & 84.4\\
			LBP \cite{OPM02} & 88.4\\
			ScatNet (NNC) \cite{BM13} & 88.6\\
			MRS4 \cite{VZ09} & 90.3\\
			SIFT + KCB \cite{CMKMV14} & 91.4\\
			MFS \cite{XJF09} & 92.7\\
			VZ-MR8 \cite{VZ05} & 92.8\\
			DeCAF \cite{CMKMV14} & 94.2\\
			FC-CNN VGGM \cite{CMKV16} & 94.5\\
			CLBP \cite{GZZ10} & 95.7\\
			\hline
			Proposed & 96.1\\
			\hline
		\end{tabular}
	\end{tabular}
	}\\
	\resizebox{.5\textwidth}{!}{
	\begin{tabular}{c}
		UMD\\			
		\begin{tabular}{cc}
			\hline
			Method & Acc. (\%)\\
			\hline
			PCANet (NNC) \cite{CJGLZM15} & 90.5\\						
			RandNet (NNC) \cite{CJGLZM15} & 90.9\\
			ScatNet (NNC) \cite{BM13} & 93.4\\			
			MFS \cite{XJF09} & 93.9\\			
           	LBP$_{riu2}$/VAR \cite{OPM02} & 95.9\\						
			LBP \cite{OPM02} & 96.1\\
			BSIF \cite{KR12} & 96.1\\
			DeCAF \cite{CMKMV14} & 96.4\\
			FC-CNN VGGM \cite{CMKV16} & 97.2\\
			SIFT + KCB \cite{CMKMV14} & 98.0\\
			SIFT+BoVW \cite{CMKMV14} & 98.1\\
			SIFT+LLC \cite{CMKV16} & 98.4\\
			CLBP \cite{GZZ10} & 98.6\\			
			\hline
			Proposed & 99.3\\
			\hline
		\end{tabular}			
	\end{tabular}
	}
\end{table}

\begin{figure}[!htpb]
	\centering
	\includegraphics[width=.5\textwidth]{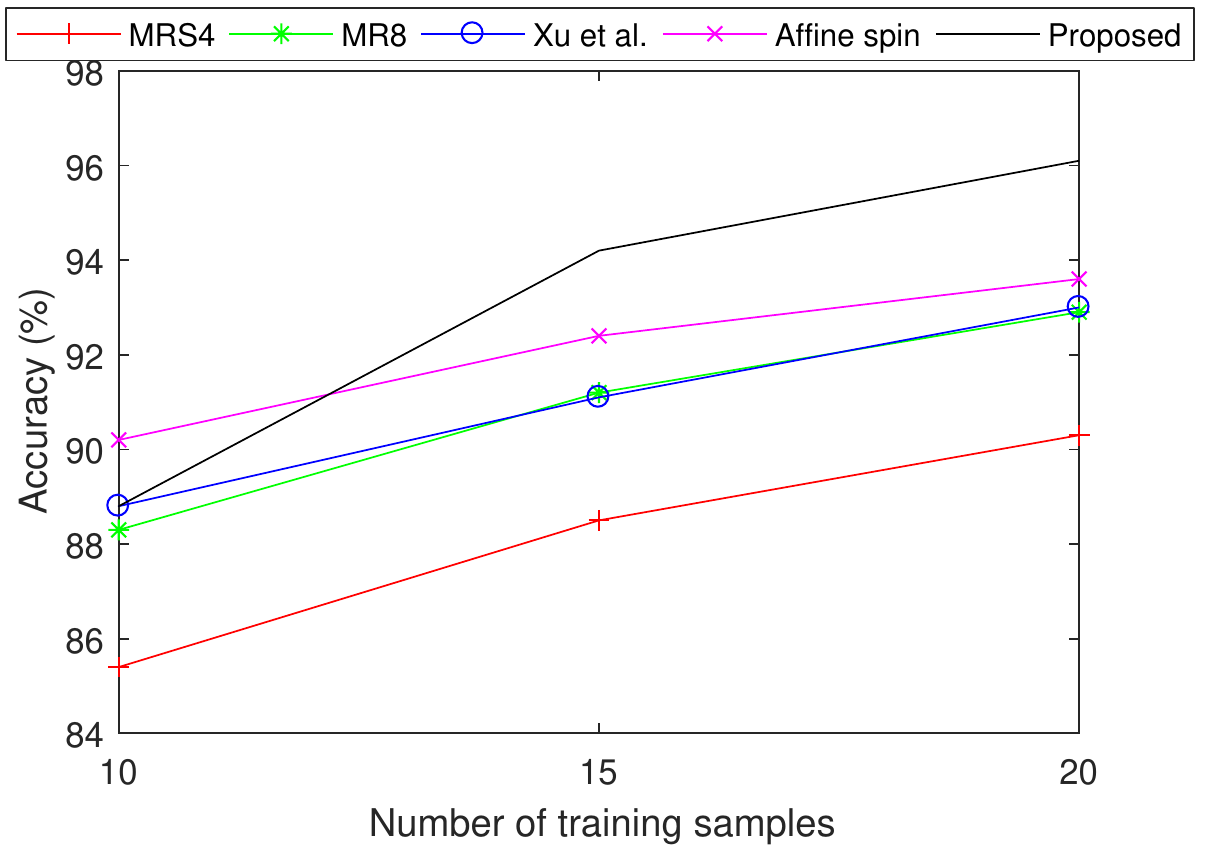}
	\caption{Accuracy when the number of training samples in UIUC database is set to 5, 10, 15, and 20.}
	\label{fig:UIUCNbrTraining}
\end{figure}

Another important test in texture recognition is on how the classifier performance is affected by a reduced number of samples for training. In particular, results on UIUC varying the number of training images are fairly common in the literature, possibly because the usual protocol (20 images for training) can be considered as not sufficiently challenging. Figure \ref{fig:UIUCNbrTraining} illustrates how the accuracy of our proposal is affected by changing the number of training samples in comparison with other methods whose corresponding results are also published in the literature. More specifically, we show the accuracy when using 10, 15 and 20 training samples and compare with MRS4 \cite{VZ09}, MR8 \cite{VZ09} and affine spin \cite{LSP05}. It can be noticed that the proposed algorithm can be considered robust to the reduction of the training subset. Its accuracy was only a little smaller than affine spin for 10 training images but was larger than the compared approaches in other situations.

\begin{figure}
	\begin{tabular}{c}
		\begin{tabular}{cc}
			\includegraphics[width=.45\textwidth]{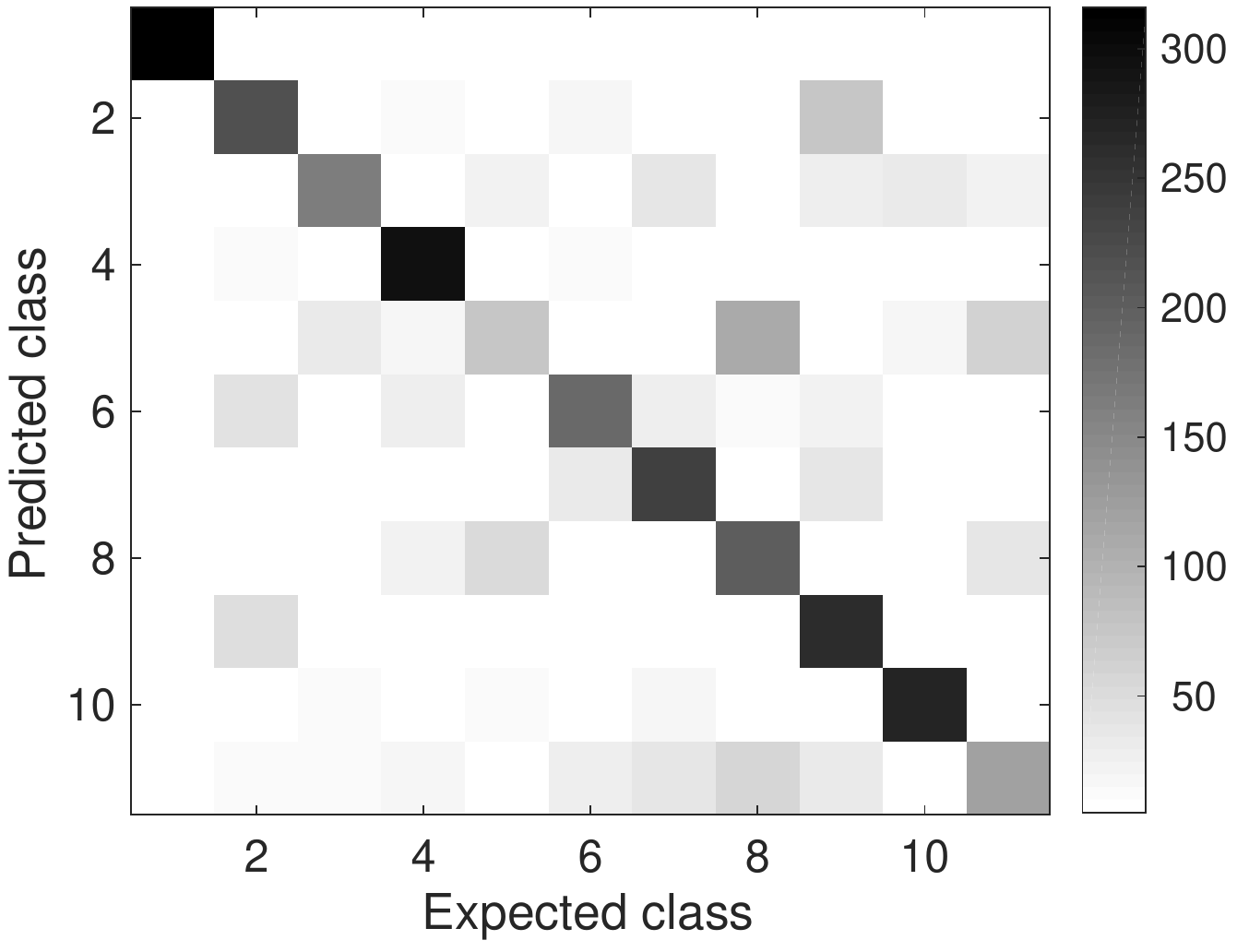} &
			\includegraphics[width=.45\textwidth]{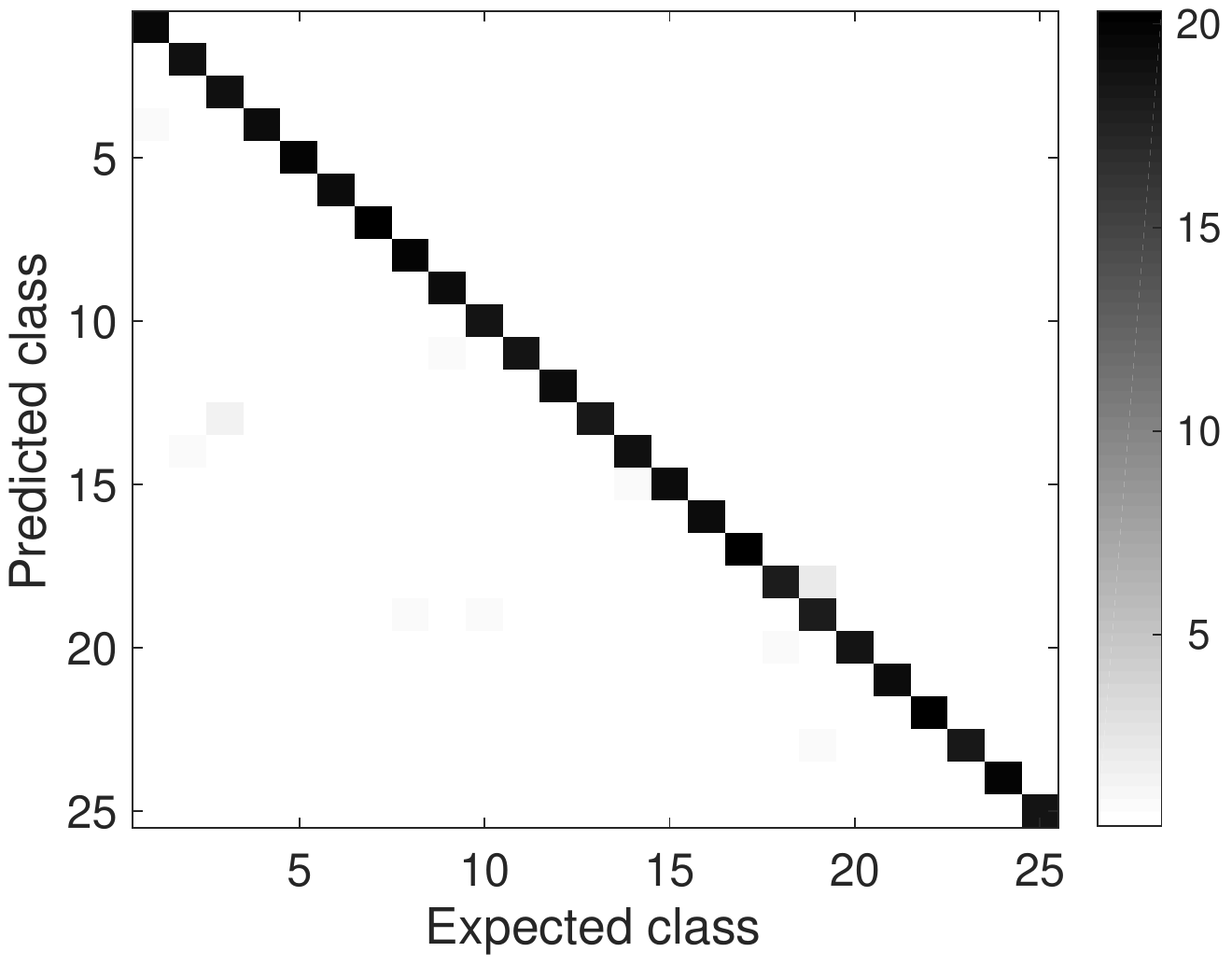}\\
			(a) & (b)\\
		\end{tabular}\\
		\begin{tabular}{c}
			\includegraphics[width=.45\textwidth]{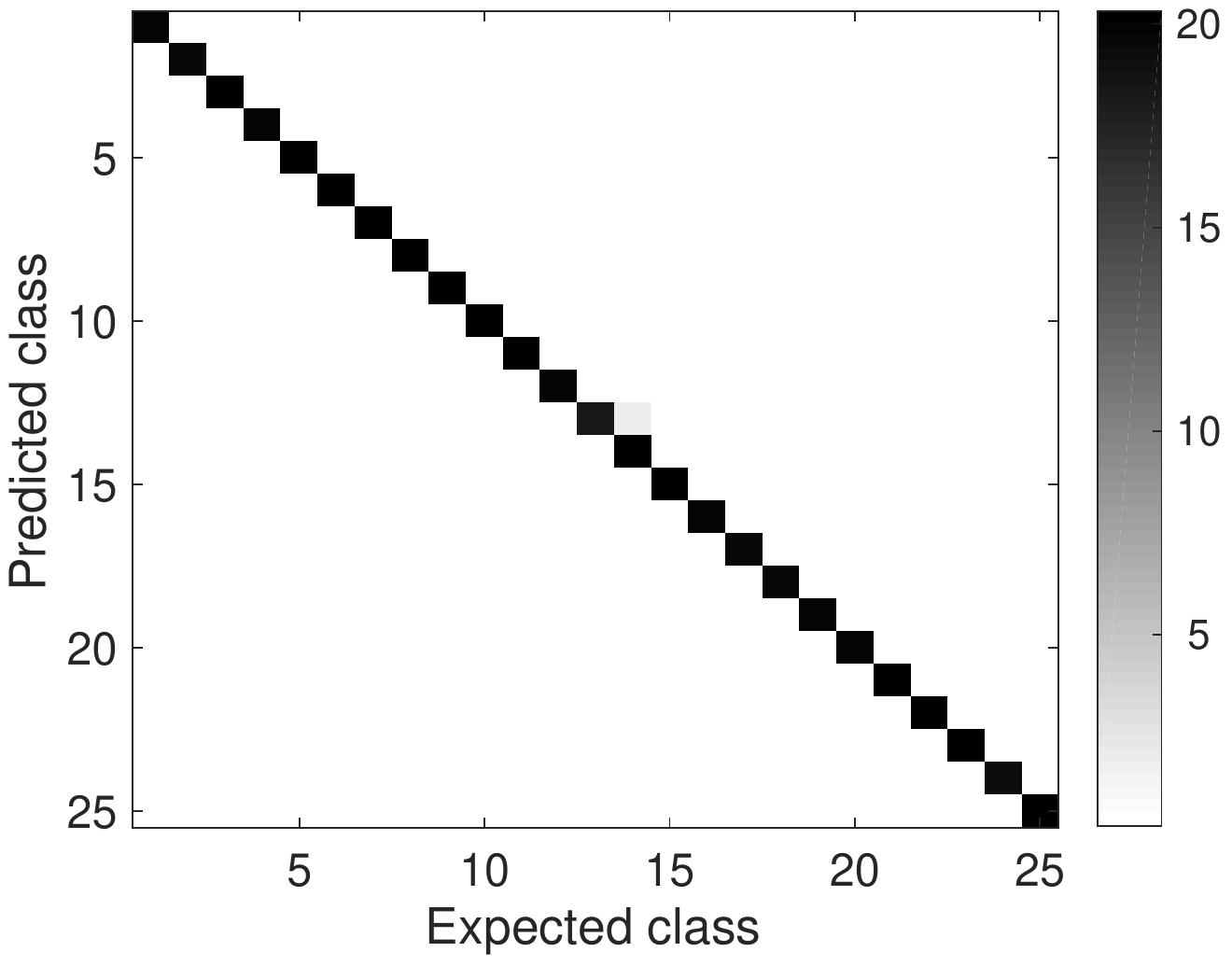}\\
			(c)\\
		\end{tabular}
	\end{tabular}
	\caption{Confusion matrices. (a) KTHTIPS-2b. (b) UIUC. (c) UMD.}
	\label{fig:CM1}
\end{figure}

Figure \ref{fig:CM1} depicts the confusion matrices for the proposed method on the benchmark datasets. In that representation, a gray map is employed to visually express the number of samples assigned by the classifier to class $A$ and actually pertaining to class $B$. An ideal matrix would present solid diagonal and no gray point outside. The pictures confirm the accuracy achieved for each database. UIUC and UMD for example has a nearly solid diagonal, as expected from their success rate in the classification close to 100\%. KTH-TIPS2b, on the other hand, has a much more complex picture. It is interesting to observe here the classes that were more difficult to classify. In this case, classes 3 (``corduroy''), 5 (``cotton''), 8 (``linen''), and 11 (``wool'') were the most usually confused. All these materials are related, as they are types of fabric, which makes such confusion somehow expected.  

In summary, the proposed methodology achieved an accuracy in the classification of benchmark texture databases that is competitive with the state-of-the-art in this topic. And this was obtained by a method that does not require neither so much computational power nor large amounts of data for training. The use of pseudo-parabolic PDE-based descriptors was expected to provide meaningful descriptors as it filters out unnecessary details and noise while effectively preserving discontinuity regions. Here, this model confirmed its capacity of providing an alternative description of the way that pixels are locally related on the image. Here such locality is quantified by the well-known LBP codes and such combination demonstrated its powerfulness in the achieved results.

\subsection{Application}

We also apply the proposed descriptors to a practical problem, to know, the identification of Brazilian plant species based on scanned images of leaf surface. The samples are collected \textit{in vivo}, washed and aligned with the basal/apical axis. The analyzed data set is called 1200Tex \cite{CMB09} and comprises a total of 1200 images, previously converted to grayscales, corresponding to 20 species (classes). Each image has dimension $128\times 128$.

Table \ref{tab:SR_1200tex} lists the accuracy for the proposed PDE descriptors on this task in comparison with other results recently published in the literature for the sabe database. It is particularly noticeable how deep learning approaches like FC-CNN and FV-CNN are outperformed by the proposed method.
\begin{table}[!htpb]
	\centering
	\caption{State-of-the-art accuracies for 1200Tex.}	
	\label{tab:SR_1200tex}
	\begin{tabular}{lc}
		\hline
		Method & Accuracy (\%)\\
		\hline
		LBPV \cite{GZZ10b} & 70.8\\
		Network diffusion \cite{GSFB16} & 75.8\\
		FC-CNN VGGM \cite{CMKV16} & 78.0\\		
		FV-CNN VGGM \cite{CMKV16} & 83.1\\		
		Gabor \cite{CMB09} & 84.0\\
		FC-CNN VGGVD \cite{CMKV16} & 84.2\\
		Schroedinger \cite{FB17} & 85.3\\		
		SIFT + BoVW \cite{CMKMV14} & 86.0\\		
		\hline
		Proposed & 87.2\\
		\hline
	\end{tabular}
\end{table}

\begin{figure}[!htpb]
	\centering
	\includegraphics[width=.45\textwidth]{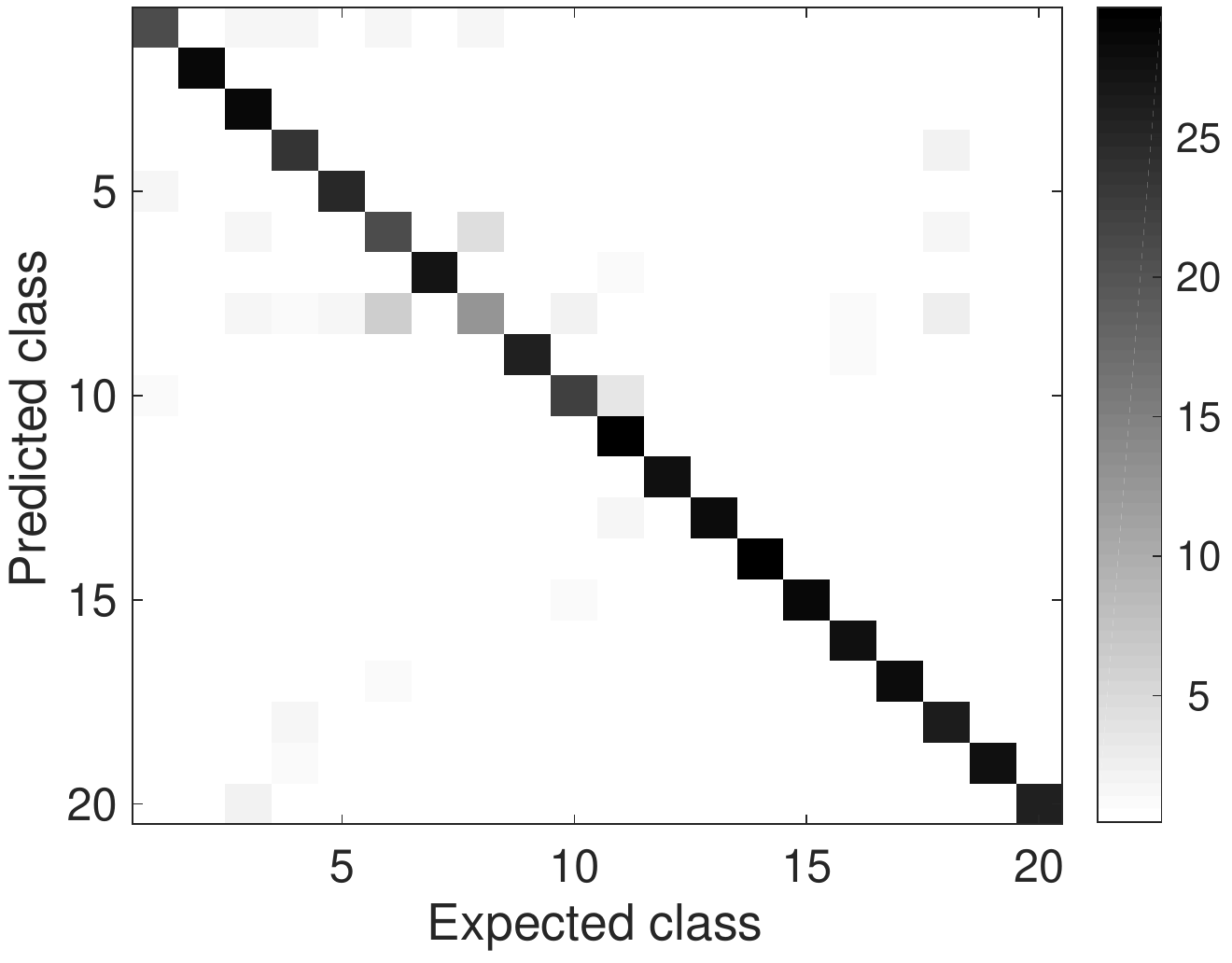}\\
	\caption{Confusion matrix for 1200Tex.}
	\label{fig:CM1200tex}
\end{figure}

\begin{figure}[!htpb]
	\centering
	\includegraphics[width=.7\textwidth]{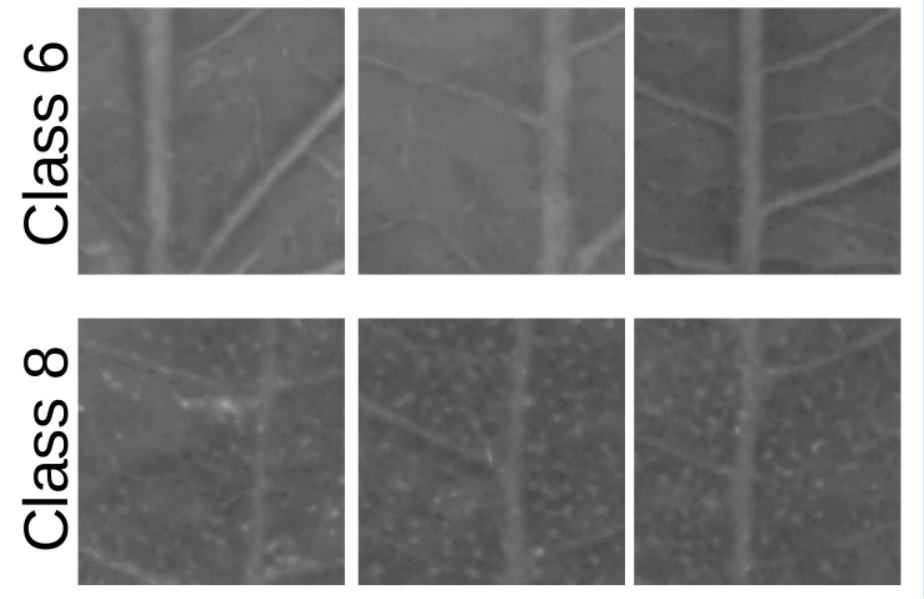}\\
	\caption{Illustrative samples of the most confused classes in 1200Tex.}
	\label{fig:1200texConfused}
\end{figure}

Figure \ref{fig:CM1200tex} depicts the corresponding confusion matrix for the proposed descriptors on 1200Tex. The matrix is nearly diagonal, as ideally expected, and the most relevant confusion takes place at class 8 (confused with class 6). Figure \ref{fig:1200texConfused} shows a few samples from those two groups illustrating the complexity of separating the respective classes. It is known in Botany that the most important elements to distinguish species when looking into leaf surface are nervures, especially primary nervure. And in Figure \ref{fig:1200texConfused} we observe that those structures are pretty similar in both classes, such that the limited performance was expected.  

In general terms, we see a good classification performance, competitive with the most advanced methods in the state-of-the-art. This is quite interesting considering that we have a model that is neither computationally expensive nor depends on large amounts of data for training. It also confirms the potential of the developed methodology in practice, especially in problems where visual texture is a relevant attribute, a common situation for example in biological applications and material sciences. 

\section{Conclusions}

This work presented a methodology to recognize texture images based on the application of a multiscale operator derived from the pseudo-parabolic diffusion PDE model.
Following the operator application, LBP codes are extracted from the transformed images to compose the feature vector, which is finally used as input to a classifier.

The method was tested on benchmark databases and in a practical application in Botany. The classification accuracy was compared to classical and state-of-the-art texture recognition approaches. The results showed that our proposal was competitive even with the most recently published approaches on this topic. This also confirmed how such ``hand-crafted'' descriptors can still be useful, especially when the computational structure at hand is not sufficiently powerful or when large amounts of data are not available, a common situation in areas like medicine, for example.

It is also important to mention that the model presented here and its effectiveness in a texture recognition task suggests the future investigation of even more complex models, such as that in \cite{abreu2017computing,AFV20}, for instance.

\section*{Conflict of interest}

The authors declare that they have no conflict of interest.

\bibliographystyle{spbasic}
\bibliography{DescPDE}

\end{document}